\documentclass{article}

\usepackage[preprint]{neurips_2026}
\makeatletter
\renewcommand{\@noticestring}{Preprint.}
\makeatother
\usepackage[utf8]{inputenc}
\usepackage[T1]{fontenc}
\usepackage{amsmath,amssymb}
\usepackage{graphicx}
\usepackage{booktabs}
\usepackage{tabularx}
\usepackage{multirow}
\usepackage{array}
\usepackage{enumitem}
\usepackage{microtype}
\usepackage{xcolor}
\usepackage{hyperref}
\hypersetup{hidelinks}

\newcolumntype{Y}{>{\raggedright\arraybackslash}X}
\newcolumntype{R}{>{\raggedleft\arraybackslash}X}

\title{Is SwiGLU's Open Positive Tail Necessary? Evidence from Closed-Tail Gating with MemGLU}
\author{%
\begin{tabular}{c@{\quad}c@{\quad}c}
\begin{tabular}[t]{c}
\textbf{Yuting Ge}$^{1,*}$\\[-1pt]
{\normalfont\scriptsize \textsuperscript{1}Department of Electrical Engineering}\\
{\normalfont\scriptsize City University of Hong Kong}\\
{\normalfont\scriptsize\texttt{yutingge3-c@my.cityu.edu.hk}}
\end{tabular}
&
\begin{tabular}[t]{c}
\textbf{Pengju Yang}$^{2,\dagger}$\\[-1pt]
{\normalfont\scriptsize \textsuperscript{2}Beihang University}\\
{\normalfont\scriptsize\texttt{yangda1223@outlook.com}}
\end{tabular}
&
\begin{tabular}[t]{c}
\textbf{Mingkai Nie}$^{3,\dagger}$\\[-1pt]
{\normalfont\scriptsize \textsuperscript{3}National University of Singapore}\\
{\normalfont\scriptsize\texttt{mingkai.nie@u.nus.edu}}
\end{tabular}%
\end{tabular}%
}

\newif\ifwithartifacts
\withartifactstrue
\ifdefined\TEXTONLY\withartifactsfalse\fi

\newcommand{\hiddenfigurelabel}[1]{\refstepcounter{figure}\label{#1}}

\begin{document}
\maketitle
\begin{center}
{\normalfont\small $^{\dagger}$ These authors contributed equally.\quad
$^*$ Corresponding author.}
\end{center}

\begin{abstract}
We test whether decoder-only language-model FFNs require SwiGLU's open positive tail. We introduce MemGLU as a closed-tail comparator derived from a memristive branch geometry. Across paired 9M and 30M pretraining runs with three seeds, MemGLU remains within about 0.1\% of SwiGLU in validation NLL. Trained SwiGLU checkpoints are sensitive to positive-tail suppression, while mechanism diagnostics show that the two models use their gates differently despite similar losses. These results suggest that models adapt to the gate geometry available during pretraining. At the tested scales, SwiGLU's open positive tail is not necessary for decoder-only language-model FFNs.
\end{abstract}


\section{Introduction}

SwiGLU~\citep{shazeer2020gluv} has become a common feed-forward design in decoder-only language models, with adoption in model families such as PaLM, LLaMA, and Qwen~\citep{chowdhery2022palm,touvron2023llama,bai2023qwen}. Its gate is
\[
\phi_{\mathrm{SiLU}}(g)=g\sigma(g).
\]
For large positive \(g\), the response remains approximately linear, leaving the positive tail open. Modern FFNs now inherit this open positive tail by default. Yet widespread adoption does not establish functional necessity. Is that tail actually necessary?

We test this question with MemGLU:
\[
\phi_{\mathrm{MemGLU}}(g)=c_0\tanh(g)\operatorname{sech}(g).
\]
MemGLU reaches a finite peak and then decays to zero on both sides, removing the open positive tail (Figure~\ref{fig:geometry}). Its shape derives from a memristor branch geometry. In the Transformer, it acts as a static elementwise gate.

We evaluate paired decoder-only language models at 9M and 30M parameters across three seeds. RMS-matched MemGLU is \(0.111\%\) better at 9M and \(0.122\%\) worse at 30M, so the direction reverses while the magnitude remains close to \(0.1\%\).

Mechanistic probes show that MemGLU models still enter the region beyond the gate peak, but this region contributes much less gate-output and GLU-product energy. Interventions on trained SwiGLU checkpoints show that suppressing the positive tail degrades performance at both scales. Together with the closely matched pretraining trajectories, these results suggest that the model adapts to the gate geometry available during training. MemGLU therefore provides a controlled counterexample to the necessity of an open positive tail at the tested scales.

Our main contributions are:

\begin{enumerate}
    \item We introduce MemGLU, a closed-tail GLU gate derived from the antisymmetric branch geometry of a first-order memristor model. To our knowledge, this is the first evaluation of this branch-derived gate in decoder-only language-model pretraining.

    \item We compare MemGLU with SwiGLU to test whether SwiGLU's open positive tail is necessary. Paired experiments at 9M and 30M yield opposite-sign differences, while the magnitude remains close to \(0.1\%\) at both scales.

    \item We identify occupancy--energy decoupling beyond the MemGLU peak and show that trained SwiGLU checkpoints are sensitive to positive-tail suppression. Together with the closely matched training trajectories, these results support adaptation to the available gate geometry during pretraining.
\end{enumerate}

\ifwithartifacts
\begin{figure}[!ht]
\centering
\includegraphics[width=\linewidth]{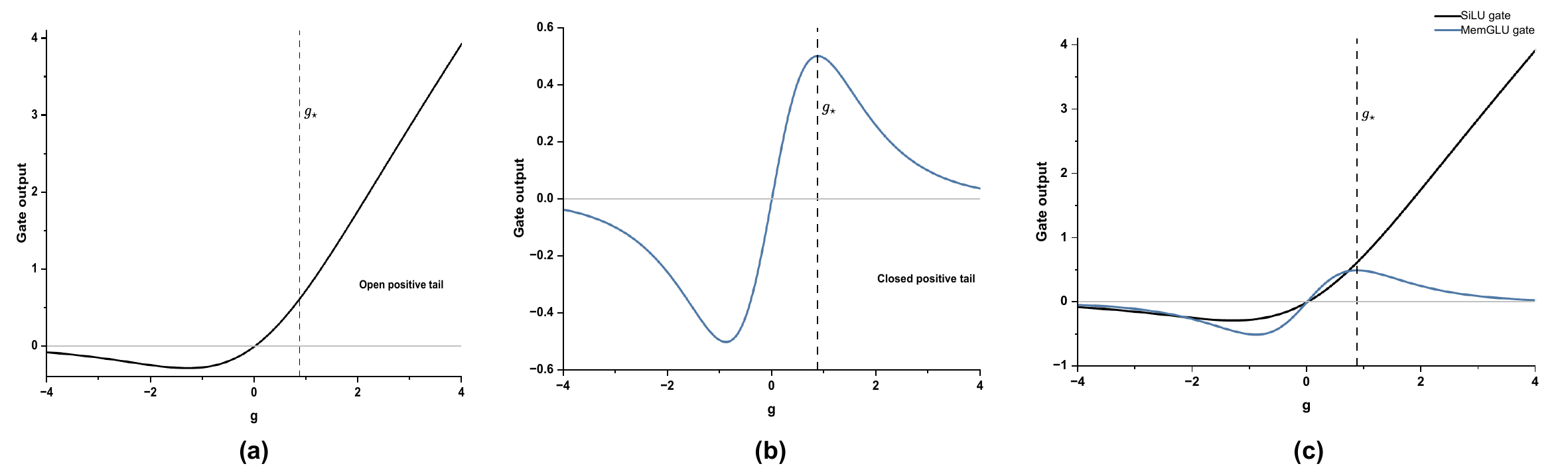}
\caption{\textbf{Gate geometry of SwiGLU and MemGLU.}
(a) The SiLU gate and its open positive tail.
(b) The MemGLU gate and its closed positive tail, with
\(g_\star=\operatorname{arsinh}(1)\) marking the positive maximum.
(c) Direct comparison of the two scalar gate responses.}
\label{fig:geometry}
\end{figure}
\else
\hiddenfigurelabel{fig:geometry}
\fi

\section{Related Work}

\subsection{Gated Feed-Forward Networks and Gate Geometry}

Gated linear units introduce multiplicative interactions by modulating a value branch with an activation-controlled gate \citep{dauphin2017glu}. Shazeer introduced several GLU variants into Transformer feed-forward networks \citep{shazeer2020gluv}. Among these variants, GEGLU and SwiGLU performed better than conventional FFNs using ReLU or GELU. SwiGLU was subsequently adopted by major decoder-only language-model families, including PaLM, LLaMA, and Qwen \citep{chowdhery2022palm,touvron2023llama,bai2023qwen}.

Recent work has examined several properties of activation and gate design. Expanded Gating Ranges directly studies the effect of extending the gate range beyond its usual interval \citep{huang2024expanded}. Other methods modify curvature, trainable shape, token adaptivity, sparsity, or numerical behavior \citep{huang2024xielu,fishman2024fp8,lee2024cats,zhuo2025polycom,jiang2026powlu,wang2026moa}. These studies develop alternative gates or improve their training properties, but they do not directly isolate whether SwiGLU's open positive tail is necessary. We test this question by comparing SwiGLU with a closed-tail gate at the same parameter count.

\subsection{Closed-Tail and Memristor-Derived Precedents}

Bounded and tail-decaying neural functions predate MemGLU. ReSech uses \(x\operatorname{sech}(x)\) as an activation in feed-forward networks \citep{njikam2016reSech}, while Fujita et al. incorporate \(\operatorname{sech}\) into recurrent gates \citep{fujita2021RNN}. These functions are mathematically related to MemGLU, but neither uses \(\tanh(g)\operatorname{sech}(g)\) as the scalar gate of a Transformer FFN. In the literature reviewed for this study, we did not identify a decoder-only language model using this exact product in a GLU.

Memristor-inspired neural activations also have established precedents. RMAF derives a static activation from a memristor-like window function \citep{yu2020rmaf}, and Li et al. implement configurable activation responses with nonlinear memristor hardware \citep{li2020memristor}. These studies apply memristor-derived nonlinearities to broader neural-network settings. In the literature reviewed here, we did not identify prior work using a memristor-derived function as the scalar GLU gate of a decoder-only language model.

\section{MemGLU: From Memristor Branch Geometry to Closed-Tail Gating}

\subsection{From Branch Separation to MemGLU}

A first order memristor model can produce two voltage branches under periodic excitation \citep{chua1971memristor,chua1976memristive}. We isolate the normalized antisymmetric separation between the two branches:

\[
\frac{v_{-}(\xi)-v_{+}(\xi)}{2\Gamma}
=
\xi\sqrt{1-\xi^{2}},
\qquad
\xi\in[-1,1].
\]

The branch separation follows the sign of the input. Its magnitude peaks at an intermediate value and returns to zero at both boundaries. These properties provide a natural closed-tail contrast to the open positive tail of SiLU. Neural gate preactivations span the real line, so we map the bounded coordinate using

\[
\xi=\tanh(g).
\]

Since

\[
\sqrt{1-\tanh^{2}(g)}
=
\operatorname{sech}(g),
\]

the branch separation becomes

\[
b(g)
=
\tanh(g)\operatorname{sech}(g).
\]

MemGLU uses the scaled gate

\[
\phi_{\mathrm{MemGLU}}(g)
=
c_{0}b(g).
\]

The unscaled response is odd and bounded. Its extrema occur at

\[
|g|
=
\operatorname{arsinh}(1),
\]

where its magnitude is \(1/2\), and it decays to zero in both directions. The full device derivation is provided in Appendix~\ref{app:derivation}.

\subsection{MemGLU in the Transformer FFN}

For an FFN input \(x\), SwiGLU computes

\[
\operatorname{FFN}_{\mathrm{SwiGLU}}(x)
=
W_{o}
\left[
\phi_{\mathrm{SiLU}}(W_{g}x)
\odot
W_{u}x
\right].
\]

MemGLU replaces only the scalar gate:

\[
\operatorname{FFN}_{\mathrm{MemGLU}}(x)
=
W_{o}
\left[
\phi_{\mathrm{MemGLU}}(W_{g}x)
\odot
W_{u}x
\right].
\]

All projection dimensions and parameter counts remain unchanged. The gate is static and elementwise, with no device state or path dependence. Figure~\ref{fig:ffn_replacement} illustrates this controlled replacement.

\ifwithartifacts
\begin{figure}[!ht]
\centering
\includegraphics[width=\linewidth]{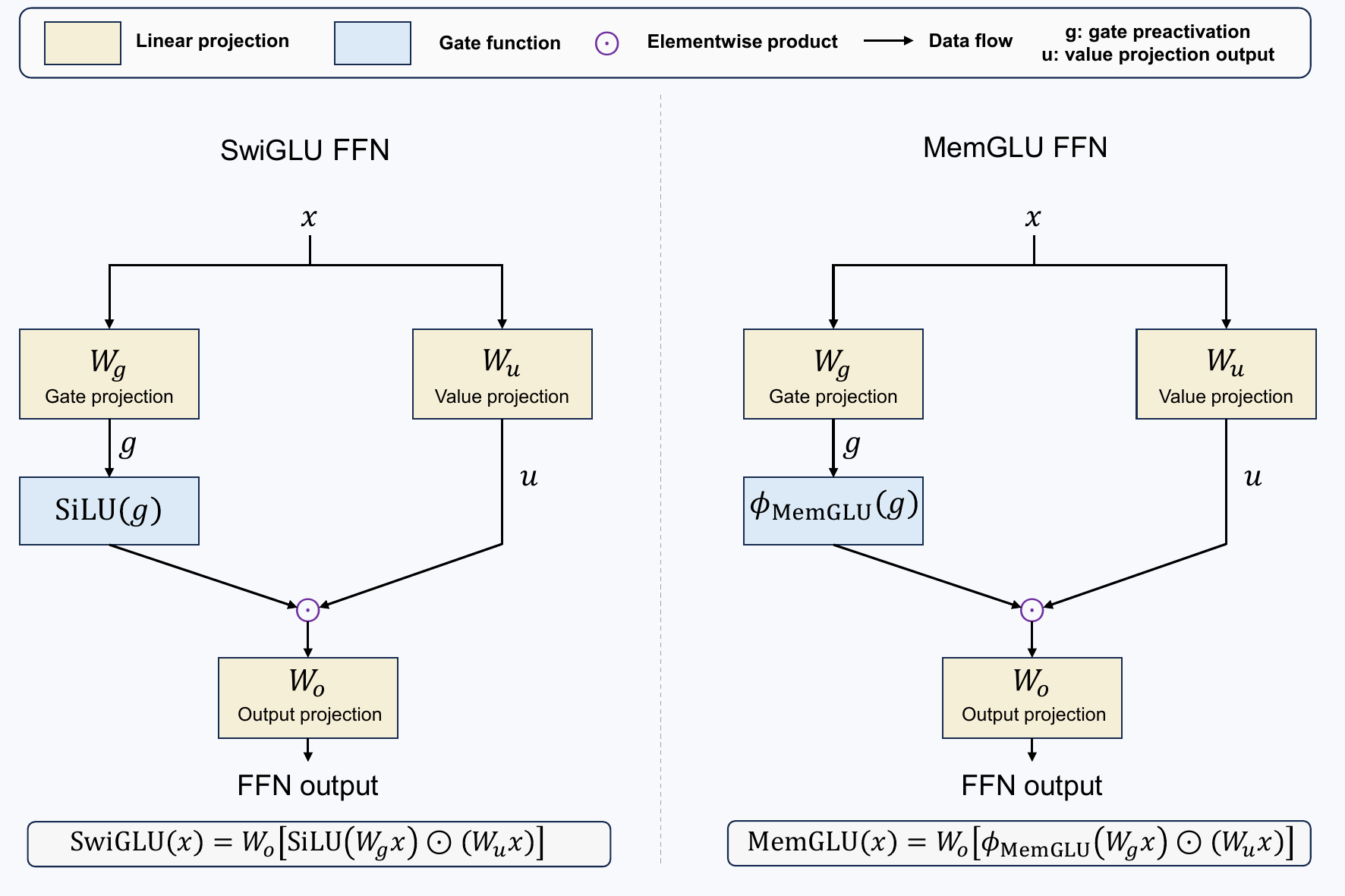}
\caption{\textbf{Controlled replacement of the scalar gate in the Transformer FFN.}
SwiGLU and MemGLU use the same projection structure and dimensions.
Only the scalar gate function is changed.}
\label{fig:ffn_replacement}
\end{figure}
\else
\hiddenfigurelabel{fig:ffn_replacement}
\fi

We evaluate an unscaled form with \(c_0=1\) and use the RMS-matched form as the primary comparison. The calibration procedure is described in Section~4.

\section{Experimental Design}

We use paired training runs to isolate the effect of the scalar gate at two language-model scales. Within each pair, SwiGLU and MemGLU share the same initial weights and data order. They also use the same token schedule, validation set, optimizer, learning rate schedule, and model dimensions. Each formal comparison uses three paired seeds.

For the RMS matched variant, \(c_{0}\) is selected to match the initialization time gate output RMS of SwiGLU:

\[
c_{0}
=
\sqrt{
\frac{
\mathbb{E}\!\left[
\phi_{\mathrm{SiLU}}(g)^{2}
\right]
}{
\mathbb{E}\!\left[
b(g)^{2}
\right]
}
}.
\]

The expectations are estimated empirically from FFN gate preactivations. The resulting \(c_0\) is shared across formal seeds and fixed throughout training.

Final validation negative log-likelihood (NLL) is the primary metric. We additionally report the mean of the final five evaluations. Complete model configurations, training budgets, and calibration details are provided in Appendix~\ref{app:configuration}.

\section{Paired Language-Model Pretraining}

We next apply the paired protocol to decoder-only language models at 9M and 30M scales.

\subsection{Training Trajectories}

At both scales, SwiGLU and RMS-matched MemGLU follow nearly indistinguishable mean training-loss and validation-NLL trajectories (Figure~\ref{fig:training_validation_curves}). The corresponding paired differences are shown in Appendix~\ref{app:paired_differences}.

\ifwithartifacts
\begin{figure}[!ht]
\centering
\includegraphics[width=0.93\linewidth]{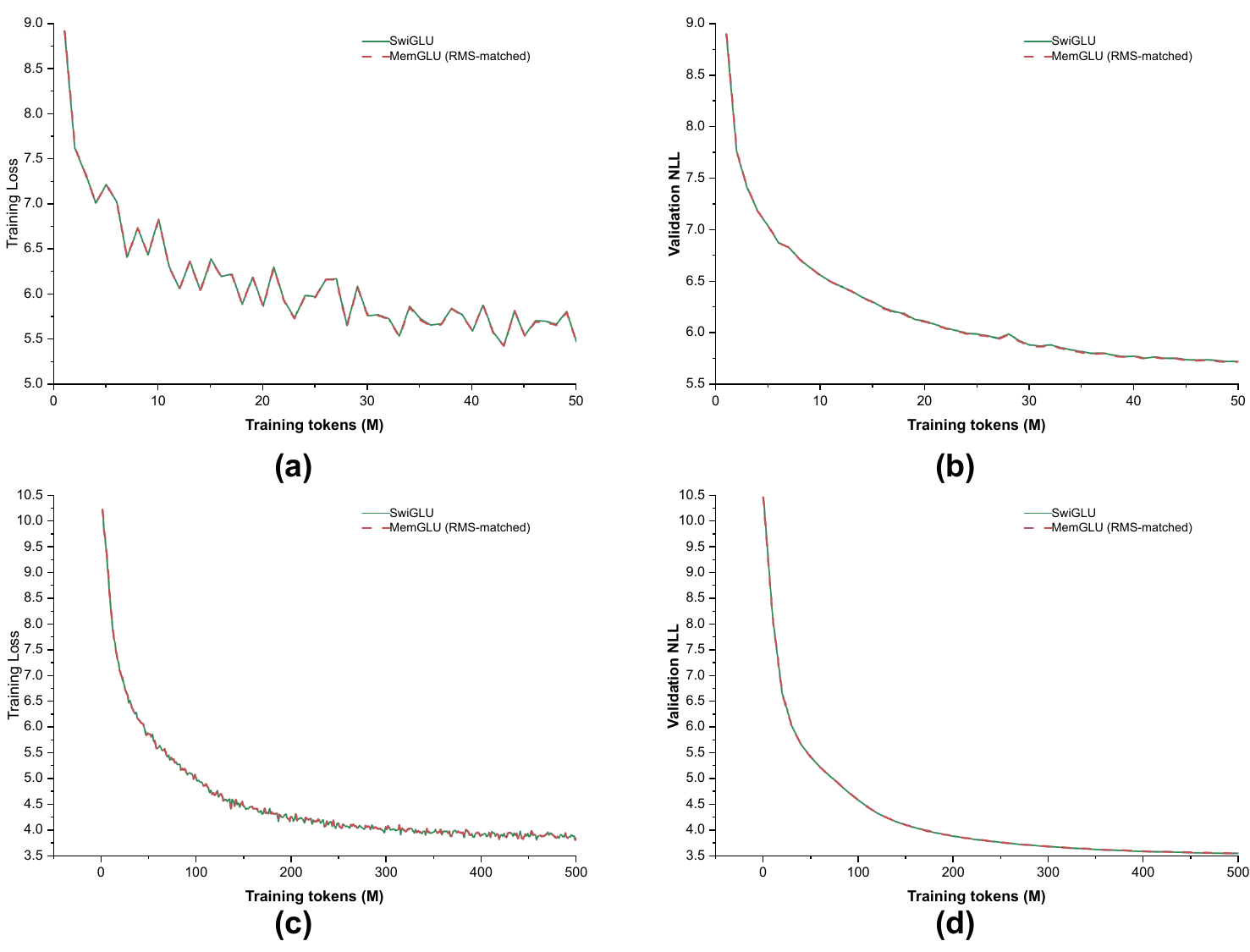}
\caption{Paired pretraining trajectories at 9M and 30M scales. (a) 9M training loss. (b) 9M validation NLL. (c) 30M training loss. (d) 30M validation NLL. Curves show three-seed means.}
\label{fig:training_validation_curves}
\end{figure}
\else
\hiddenfigurelabel{fig:training_validation_curves}
\fi

\subsection{Final Paired Comparison}

\begin{table*}[!ht]
\centering
\small
\setlength{\tabcolsep}{3.5pt}
\renewcommand{\arraystretch}{1.28}
\caption{Paired language-modeling results across three seeds. Values are mean $\pm$ sample standard deviation. Relative final differences use SwiGLU at the same scale as the reference. Lower NLL is better. The unscaled 9M variant is an auxiliary unscaled comparison. The RMS-matched variant is the primary comparison.}
\label{tab:paired_lm_results}
\newcommand{\meanpm}[2]{\shortstack[c]{\(\displaystyle #1\)\\[0.45ex]\(\pm\,#2\)}}
\begin{tabularx}{\textwidth}{@{}cYcccc@{}}
\toprule
Scale & Gate & \shortstack[c]{Final NLL\\$\downarrow$} & \shortstack[c]{Final-five NLL\\$\downarrow$} & \shortstack[c]{Relative final\\$\Delta$} & \shortstack[c]{Paired\\wins} \\
\midrule
\multirow{3}{*}{9M}
 & SwiGLU & \meanpm{5.726682}{0.007679} & \meanpm{5.732096}{0.007528} & \textit{reference} & -- \\
\addlinespace[0.4em]
 & MemGLU (RMS-matched) & \meanpm{5.720326}{0.002360} & \meanpm{5.725963}{0.002470} & $-0.111\%$ & 3/3 \\
\addlinespace[0.4em]
 & MemGLU (unscaled) & \meanpm{5.696731}{0.010864} & \meanpm{5.702339}{0.011563} & $-0.523\%$ & 3/3 \\
\addlinespace[0.65em]
\multirow{2}{*}{30M}
 & SwiGLU & \meanpm{3.553958}{0.000169} & \meanpm{3.557468}{0.000179} & \textit{reference} & -- \\
\addlinespace[0.4em]
 & MemGLU (RMS-matched) & \meanpm{3.558291}{0.003820} & \meanpm{3.561710}{0.003779} & $+0.122\%$ & 0/3 \\
\bottomrule
\end{tabularx}
\end{table*}

Table 1 confirms the late-stage pattern. At 9M scale, RMS-matched MemGLU achieves lower final NLL in all three paired seeds, with a mean relative difference of $-0.111\%$. At 30M scale, its final NLL is higher in all three paired seeds, with a difference of $+0.122\%$. Across the two scales, the sign reverses while the magnitude remains close to $0.1\%$.

\section{Gate Usage and Positive Tail Sensitivity}

At the tested scales, a decoder-only language model can be trained to near-SwiGLU performance with a closed-tail gate. The next question is how MemGLU operates inside the trained FFN.

\subsection{Occupancy and Energy Beyond the MemGLU Peak}

\ifwithartifacts
\begin{figure}[!ht]
\centering
\includegraphics[width=\linewidth]{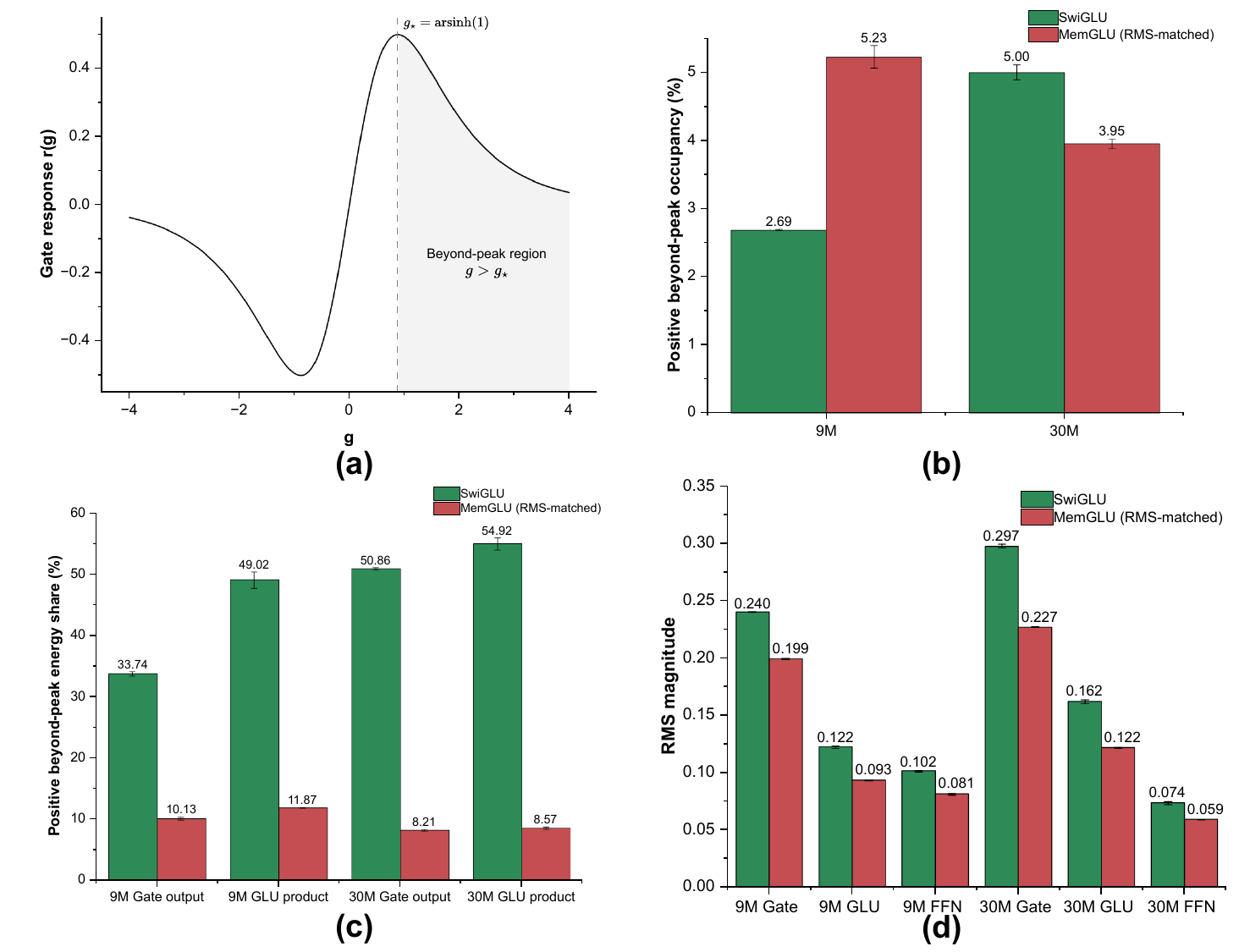}
\caption{Occupancy--energy decoupling beyond the MemGLU peak. (a) The region \(g>g_\star\). (b) Positive beyond-peak occupancy. (c) Gate output and GLU product energy shares from \(g>g_\star\). (d) RMS magnitudes of the gate output, GLU product, and FFN update.}
\label{fig:mechanism}
\end{figure}
\else
\hiddenfigurelabel{fig:mechanism}
\fi

MemGLU reaches its positive maximum at
\[
g_\star=\operatorname{arsinh}(1).
\]
We define \(g>g_\star\) as the positive beyond-peak region shown in Figure~\ref{fig:mechanism}a. A simple explanation is that MemGLU learns to avoid the descending part of its gate. Figure~\ref{fig:mechanism}b rules out this explanation through the occupancy measurements.

At 9M, positive beyond-peak occupancy increases from $2.688\%$ under SwiGLU to $5.230\%$ under RMS-matched MemGLU. At 30M, it decreases from $5.002\%$ under SwiGLU to $3.954\%$ under RMS-matched MemGLU. The direction of change reverses across scales.

Figure~\ref{fig:mechanism}c shows the energy measurements. At 9M, the gate-output energy share contributed by the positive beyond-peak region falls from $33.736\%$ under SwiGLU to $10.129\%$ under MemGLU. The corresponding GLU-product energy share falls from $49.018\%$ to $11.870\%$. At 30M, the gate-output share falls from $50.857\%$ to $8.209\%$, and the GLU-product share falls from $54.921\%$ to $8.573\%$. The region can remain populated while contributing a much smaller fraction of the squared energy in the gated branch. We refer to this separation as \emph{occupancy--energy decoupling}.

Figure~\ref{fig:mechanism}d shows lower RMS magnitudes for the gate output, GLU product, and FFN update under MemGLU. Appendix~\ref{app:mechanism_metrics} provides the full RMS ratios, including the gate preactivation \(g\), together with metric definitions and aggregation details.

\subsection{Positive Tail Sensitivity in Trained SwiGLU}

We apply hard removal, positive capping, and smooth attenuation to trained SwiGLU checkpoints only during evaluation (Figure~\ref{fig:tail_interventions}). Exact definitions are provided in Appendix~\ref{app:intervention_definitions}.

\ifwithartifacts
\begin{figure}[!ht]
    \centering
    \includegraphics[width=\linewidth]{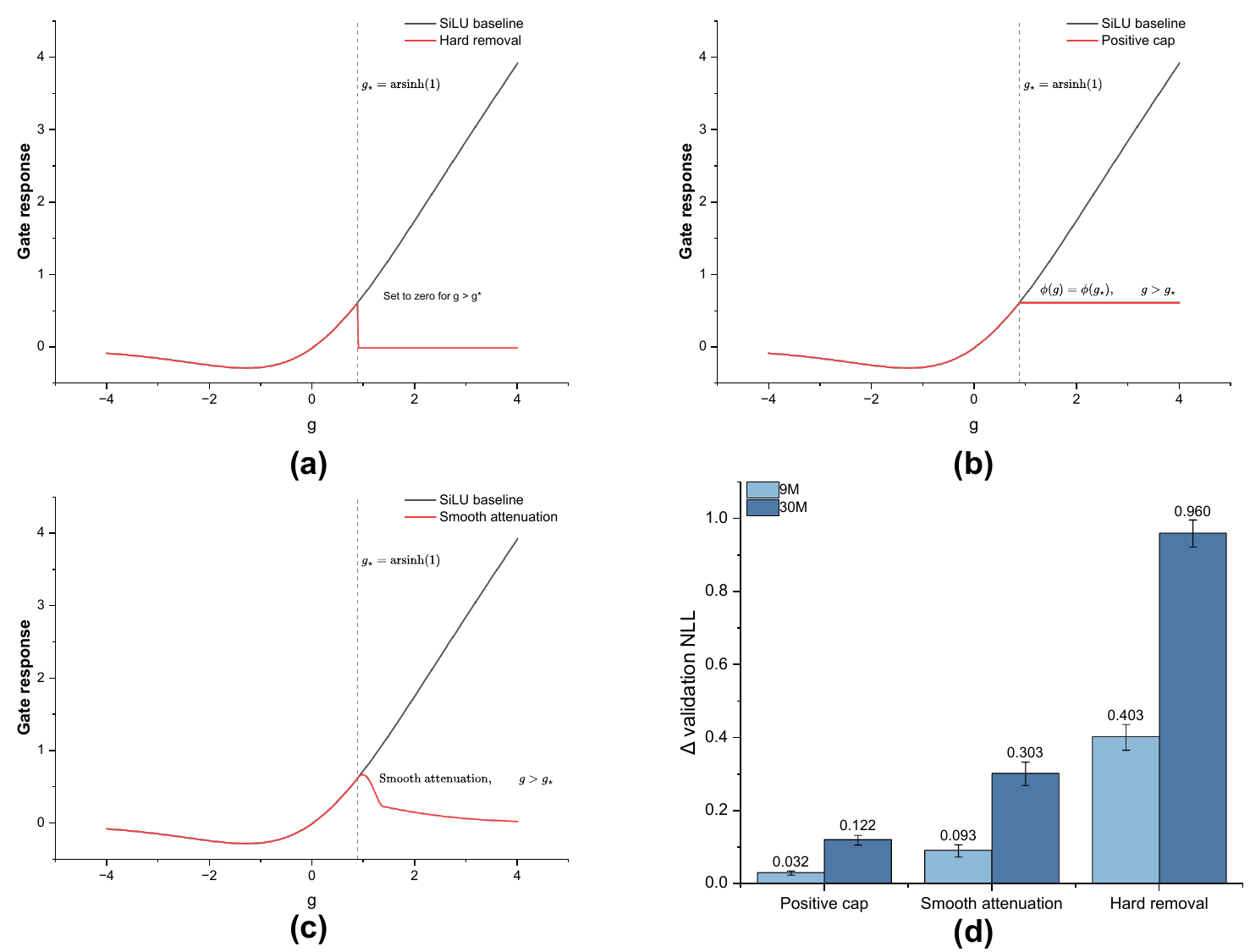}
    \caption{Positive-tail interventions on trained SwiGLU checkpoints. (a) Hard removal. (b) Positive cap. (c) Smooth attenuation. (d) Validation-NLL increase at 9M and 30M scales.}
    \label{fig:tail_interventions}
\end{figure}
\else
\hiddenfigurelabel{fig:tail_interventions}
\fi

Every intervention increases validation NLL in all three seeds at both scales. The ordering is identical across the two model sizes. Positive capping produces the smallest increase, smooth attenuation produces an intermediate increase, and hard removal produces the largest increase. At 9M, the mean $\Delta\mathrm{NLL}$ values are $0.03173$, $0.09258$, and $0.40291$, respectively. At 30M, they increase to $0.12194$, $0.30288$, and $0.95986$. Every intervention has a larger effect at 30M.

\section{Discussion and Limitations}

\subsection{Can the Model Adapt Without an Open Tail?}

Figure~\ref{fig:training_validation_curves} and Table~\ref{tab:paired_lm_results} show that SwiGLU and MemGLU follow closely matched pretraining trajectories and reach similar language-modeling performance. A model trained from the start with a closed positive tail can therefore reach near-SwiGLU performance at both tested scales. This provides a counterexample to the necessity of SwiGLU's open positive tail.

At the same time, Figure~\ref{fig:tail_interventions} shows that trained SwiGLU checkpoints are sensitive to positive-tail suppression after pretraining. Figure~\ref{fig:mechanism} shows that MemGLU models continue to enter the beyond-peak region, but that region contributes much less gate-output and GLU-product energy than under SwiGLU. Together, these results show that the two models reach similar performance while using their gates differently.

But why can gate geometry shape how a trained model uses its FFN without substantially limiting the performance it reaches? Gate geometry is one of the fixed constraints under which optimization proceeds, and every parameter update occurs with that geometry in place. Training under different gate geometries can therefore produce different FFN usage patterns while still reaching similar losses. This is consistent with an adaptation account: pretraining forms an FFN usage pattern compatible with the gate geometry available throughout training. At the tested scales, gate geometry therefore appears to shape how the trained FFN operates without strongly constraining the final language-modeling loss.

\subsection{Scope and Limitations}

The evidence is limited to 9M and 30M decoder-only models, with three paired seeds at each scale and a shared training setup. The proposed adaptation account is inferred from the close pretraining trajectories and the final checkpoint diagnostics. Broader validation will require experiments with substantially larger models.

The current MemGLU implementation does not use a fused kernel. In the measured 30M setup, it reduces training throughput by \(10.46\%\) and increases peak memory by \(18.81\%\). Larger models, broader training settings, and optimized kernels are needed to determine how far the results generalize.

\section{Conclusion}

We asked whether SwiGLU's open positive tail is necessary in decoder-only language-model FFNs. To test this, we replaced the SiLU gate with MemGLU, whose response closes on the positive side. Across paired 9M and 30M pretraining runs, MemGLU remains within about 0.1\% of SwiGLU. The diagnostic and intervention results further indicate that similar final losses can accompany different use of gate geometry. This suggests that pretraining adapts to the gate geometry available throughout training. At the tested scales, SwiGLU's open positive tail is not necessary for decoder-only language-model FFNs.

\bibliographystyle{plainnat}
\bibliography{references}

\clearpage
\appendix
\small

\section{Full Derivation of the Memristive Branch Geometry}
\label{app:derivation}

Following the first-order memristor formulation \citep{chua1971memristor,chua1976memristive}, we consider

\[
v=R(w)i,\qquad
\dot w=\kappa i,\qquad
R(w)=R_0+\beta(w-w_0).
\]

Here, \(v\) and \(i\) denote the device voltage and current. The internal state variable is \(w\), with \(\dot w=\kappa i\) specifying its evolution. The resistance \(R(w)\) varies linearly with the state around the reference value \(w_0\). \(R_0\) is the corresponding reference resistance, and \(\beta\) controls the state dependence.

Under the sinusoidal input

\[
i(t)=I\sin(\omega t),
\]

the state equation becomes

\[
\dot w=\kappa I\sin(\omega t).
\]

Integrating with respect to time gives

\[
w(t)-w_0
=
-\frac{\kappa I}{\omega}\cos(\omega t),
\]

where the integration constant is absorbed into the reference state \(w_0\).

We now define the normalized input coordinate

\[
\xi=\sin(\omega t)\in[-1,1].
\]

For a fixed value of \(\xi\), there are two corresponding phases of the sinusoidal cycle with opposite cosine signs:

\[
\cos(\omega t)
=
\pm\sqrt{1-\xi^2}.
\]

Substituting these two possibilities into the state expression gives the two corresponding state branches,

\[
w_\pm(\xi)-w_0
=
\mp
\frac{\kappa I}{\omega}
\sqrt{1-\xi^2}.
\]

Using \(R(w)=R_0+\beta(w-w_0)\), these state branches produce two resistance branches,

\[
R_\pm(\xi)
=
R_0
\mp
\frac{\beta\kappa I}{\omega}
\sqrt{1-\xi^2}.
\]

The current at the same normalized coordinate is \(i=I\xi\). Substituting the resistance branches into \(v=R(w)i\) therefore gives

\[
v_\pm(\xi)
=
R_0I\xi
\mp
\frac{\beta\kappa I^2}{\omega}
\xi\sqrt{1-\xi^2}.
\]

Defining

\[
\Gamma=\frac{\beta\kappa I^2}{\omega},
\]

we obtain

\[
v_\pm(\xi)
=
R_0I\xi
\mp
\Gamma\xi\sqrt{1-\xi^2}.
\]

The two branches share the same term \(R_0I\xi\). Their difference cancels this common component and isolates the branch-dependent part:

\[
v_-(\xi)-v_+(\xi)
=
2\Gamma\xi\sqrt{1-\xi^2}.
\]

Normalizing by \(2\Gamma\) gives the antisymmetric branch separation

\[
\frac{v_-(\xi)-v_+(\xi)}{2\Gamma}
=
\xi\sqrt{1-\xi^2}.
\]

For neural gating, we map the bounded coordinate \(\xi\in[-1,1]\) to the real-valued gate preactivation \(g\) using

\[
\xi=\tanh(g).
\]

Since

\[
\sqrt{1-\tanh^2(g)}
=
\operatorname{sech}(g),
\]

the normalized branch separation becomes

\[
b(g)
=
\tanh(g)\operatorname{sech}(g),
\]

and the MemGLU gate is

\[
\phi_{\mathrm{MemGLU}}(g)
=
c_0 b(g).
\]

For the scalar basis

\[
r(z)=\tanh(z)\operatorname{sech}(z),
\]

its derivative is

\[
r'(z)
=
\operatorname{sech}(z)
\left(1-2\tanh^2(z)\right).
\]

Its extrema occur at

\[
z=\pm\operatorname{arsinh}(1),
\qquad
\left|r\!\left(\pm\operatorname{arsinh}(1)\right)\right|
=
\frac12,
\]

and near the origin,

\[
r(z)
=
z-\frac56z^3+O(z^5).
\]

Moreover,

\[
\lim_{z\to\pm\infty}r(z)=0.
\]

MemGLU comes from a normalized antisymmetric branch separation. The relation \(\xi=\tanh(g)\) is a design mapping for neural gating, and in a Transformer the resulting gate is a static elementwise function.

\clearpage
\section{Experimental Configurations}
\label{app:configuration}
\setcounter{table}{0}
\renewcommand{\thetable}{\thesection\arabic{table}}
\renewcommand{\theHtable}{\thesection\arabic{table}}

\begin{table}[!ht]
\centering
\caption{Model and training configurations for the paired 9M and 30M experiments.}
\label{tab:appendix_config}
\begin{tabularx}{\textwidth}{@{}l>{\centering\arraybackslash}X>{\centering\arraybackslash}X@{}}
\toprule
Configuration & 9M & 30M \\
\midrule
Parameter count & 8,948,160 & 30,285,696 \\
Number of layers & 6 & 10 \\
Model width ($d_{\mathrm{model}}$) & 192 & 384 \\
Attention heads & 3 & 6 \\
FFN width ($d_{\mathrm{ff}}$) & 512 & 1,024 \\
Context length & 256 & 512 \\
Formal training tokens/run & 50,000,000 & 499,908,608 \\
Training steps & 3,052 & 3,814 \\
Paired seeds & 42, 43, 44 & 42, 43, 44 \\
Optimizer & AdamW & fused AdamW \\
Base learning rate & $3.0\times10^{-4}$ & $3.0\times10^{-4}$ \\
RMS calibration $c_0$ & 0.6039096539628113 & 0.6990967290796867 \\
\bottomrule
\end{tabularx}
\end{table}

\noindent\textbf{Data.} Both scales use the same pre-tokenized corpus with disjoint training and validation shards. The vocabulary size is 32,768.

\noindent\textbf{Optimization.} The 9M and 30M runs use effective batches of 16,384 and 131,072 target tokens per update, respectively. Both use AdamW with a learning rate of \(3\times10^{-4}\), betas \((0.9,0.95)\), weight decay \(0.1\), gradient clipping at \(1.0\), and no dropout. The learning rate follows cosine decay to \(10\%\) of its initial value. Warmup covers \(2\%\) of training at 9M and \(4\%\) at 30M. The 9M runs use FP16 autocast, while the 30M runs use BF16.

\noindent\textbf{Logging and evaluation.} At 9M, training loss and validation NLL are recorded at approximately 1M-token intervals. At 30M, training loss is logged every 10 optimizer steps, and validation NLL is evaluated at approximately 10M-token intervals.

\noindent\textbf{RMS calibration.} The scale-specific \(c_0\) is estimated before formal training from FFN gate preactivations. At each scale, training-token windows are passed through 32 randomly initialized models and the resulting preactivations are collected for calibration. The resulting \(c_0\) is shared across the three formal seeds at that scale and fixed throughout training. The values used in the experiments are reported in Table B1.

\noindent\textbf{Diagnostics.} Each mechanism and intervention diagnostic uses 1,048,576 validation tokens.

\clearpage
\section{Paired Training and Validation Differences}
\label{app:paired_differences}
\setcounter{figure}{0}
\renewcommand{\thefigure}{\thesection\arabic{figure}}
\renewcommand{\theHfigure}{\thesection\arabic{figure}}

Figure~\ref{fig:paired_loss_differences} reports the paired differences corresponding to Figure~\ref{fig:training_validation_curves}.

\ifwithartifacts
\begin{figure}[!ht]
    \centering
    \includegraphics[width=\linewidth]{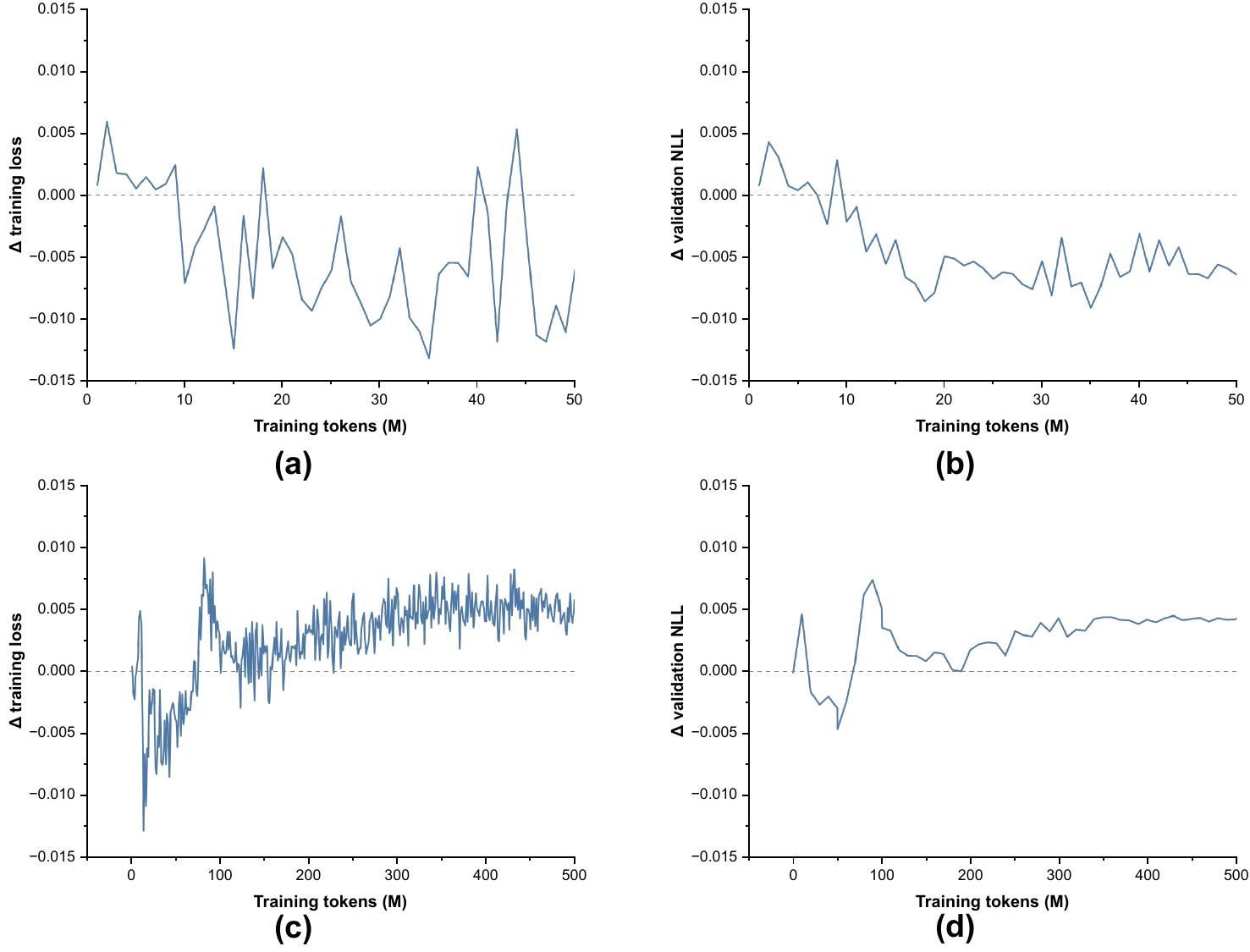}
    \caption{Mean paired differences corresponding to Figure~\ref{fig:training_validation_curves}. \(\Delta=L_{\mathrm{MemGLU}}-L_{\mathrm{SwiGLU}}\). Panels (a)--(d) follow the order of Figure~\ref{fig:training_validation_curves}. Negative values indicate lower MemGLU loss, and positive values indicate lower SwiGLU loss.}
    \label{fig:paired_loss_differences}
\end{figure}
\else
\hiddenfigurelabel{fig:paired_loss_differences}
\fi

\clearpage
\section{Mechanism Metrics and RMS Propagation}
\label{app:mechanism_metrics}

\setcounter{table}{0}
\renewcommand{\thetable}{\thesection\arabic{table}}
\renewcommand{\theHtable}{\thesection\arabic{table}}

\noindent\textbf{Metric definitions.} For an FFN input \(x\), we record the following quantities:

\[
g=W_gx,\qquad
q=\phi(g),\qquad
u=W_ux,
\]

\[
h=q\odot u,\qquad
\Delta x=W_oh.
\]

Here, \(g\) is the gate preactivation, \(q\) is the gate output, \(u\) is the value-branch output, \(h\) is the GLU product, and \(\Delta x\) is the FFN update.

We use

\[
g_\star=\operatorname{arsinh}(1)
\]

to define the positive beyond-peak region. For each gate preactivation element \(g_i\),

\[
m_i=\mathbf{1}[g_i>g_\star].
\]

The positive beyond-peak occupancy is the fraction of gate preactivation elements that fall in this region:

\[
\mathrm{Occ}
=
\frac{\sum_i m_i}{\sum_i 1}.
\]

To measure how much squared magnitude comes from the same region, we define the gate-output energy share as

\[
S_q
=
\frac{\sum_i m_iq_i^2}
{\sum_i q_i^2},
\]

and the GLU-product energy share as

\[
S_h
=
\frac{\sum_i m_ih_i^2}
{\sum_i h_i^2}.
\]

For these three metrics, the index \(i\) runs over all layers, diagnostic batches, token positions, and FFN channels in the fixed validation probe.

\noindent\textbf{RMS magnitudes.} We also measure the global RMS magnitude of the gate preactivation, gate output, GLU product, and FFN update. For each quantity \(z\),

\[
\operatorname{RMS}(z)
=
\sqrt{
\frac{1}{N_z}
\sum_i z_i^2
},
\qquad
z\in\{g,q,h,\Delta x\},
\]

where \(N_z\) is the total number of scalar elements accumulated for \(z\).

For each seed and gate type, squared quantities are accumulated over the full diagnostic probe before the RMS is computed. The values in Figure 4d are therefore global RMS magnitudes across the diagnostic set.

\noindent\textbf{Across-seed summary.} Each reported mechanism metric is computed separately for the three seeds. Figure 4 reports the mean across seeds, with error bars showing the sample standard deviation.

Table D1 reports the ratio

\[
\frac{\operatorname{RMS}_{\mathrm{MemGLU}}}
{\operatorname{RMS}_{\mathrm{SwiGLU}}}
\]

for \(g\), \(q\), \(h\), and \(\Delta x\). Values below \(1\) indicate a smaller RMS magnitude under MemGLU.

\begin{table}[!ht]
\centering
\caption{RMS ratios for MemGLU relative to SwiGLU. Values below 1 indicate smaller RMS under MemGLU.}
\label{tab:rms_ratios}
\begin{tabular}{lcccc}
\toprule
Scale & $g$ & $q$ & $h$ & $\Delta x$ \\
\midrule
9M & 1.209 & 0.830 & 0.763 & 0.802 \\
30M & 0.949 & 0.763 & 0.751 & 0.803 \\
\bottomrule
\end{tabular}
\end{table}

\clearpage
\section{Positive-Tail Intervention Definitions}
\label{app:intervention_definitions}

All interventions use the threshold \(g_\star=\operatorname{arsinh}(1)\). This is the positive peak of the MemGLU gate and marks where modification of the SwiGLU positive tail begins.

The baseline gate is
\[
\phi_{\mathrm{SiLU}}(g)=g\sigma(g).
\]
Hard removal and positive capping are defined by
\[
\phi_{\mathrm{hard}}(g)=
\begin{cases}
\phi_{\mathrm{SiLU}}(g), & g\le g_\star,\\
0, & g>g_\star,
\end{cases}
\qquad
\phi_{\mathrm{cap}}(g)=
\begin{cases}
\phi_{\mathrm{SiLU}}(g), & g\le g_\star,\\
g_\star\sigma(g_\star), & g>g_\star.
\end{cases}
\]
For smooth attenuation, we first define the SiLU value at the threshold

\[
c_\star=g_\star\sigma(g_\star).
\]

We use a transition width \(w=0.5\) and decay parameter \(\beta=1.0\). For \(g>g_\star\), the normalized distance from the threshold is

\[
t(g)=
\operatorname{clip}
\left(
\frac{g-g_\star}{w},
0,1
\right).
\]

A smooth interpolation weight is then defined as

\[
m(g)=t(g)^2\left(3-2t(g)\right).
\]

The attenuated gate is

\[
\phi_{\mathrm{smooth}}(g)=
\begin{cases}
\phi_{\mathrm{SiLU}}(g), & g\le g_\star,\\
\begin{aligned}
&\bigl(1-m(g)\bigr)\phi_{\mathrm{SiLU}}(g)\\
&\quad +m(g)c_\star\exp\!\left[-\beta\operatorname{softplus}(g-g_\star)\right],
\end{aligned}
& g>g_\star,
\end{cases}
\]

Here,

\[
\operatorname{softplus}(z)=\log(1+e^z).
\]
All interventions modify only the forward gate during evaluation and perform no retraining or parameter updates.

\normalsize

\end{document}